\documentclass[10pt,twocolumn,letterpaper]{article}

\usepackage{cvpr}
\usepackage{times}
\usepackage{epsfig}
\usepackage{graphicx}
\usepackage{amsmath}
\usepackage{amssymb}
\usepackage{microtype}
\usepackage{booktabs}
\usepackage{threeparttable}
\usepackage{mathtools}
\usepackage{tabularx}
\usepackage{enumitem}

\cvprfinalcopy %

\def\cvprPaperID{2709} %

\ifcvprfinal\fi
\begin{document}

\title{Single-View View Synthesis with Multiplane Images}

\author{Richard Tucker\hspace{3em}Noah Snavely\\
Google Research\\
{\tt\small richardt@google.com\hspace{2em}snavely@google.com} \\
}

\maketitle

\thispagestyle{empty}

\frenchspacing

\def\Image{\mathbf{I}}
\def\Imagehat{\hat{\mathbf{I}}}
\def\Disparity{\hat{\mathbf{D}}}
\def\Points{\mathbf{P}}
\def\MPI{MPI}
\def\dnear{d_\textsf{near}}
\def\dfar{d_\textsf{far}}
\def\scale{\sigma}
\def\Network{f}
\def\Render{\mathcal{R}}
\def\M{\mathbf{M}}
\def\Sobel{\mathbf{G}}
\def\emin{e_\textsf{min}}
\def\gmin{g_\textsf{min}}
\def\blendsteps{s_\textsf{bg}}
\def\down{$\textrm{MP}_2$}
\def\up{$\textrm{Up}_2$}
\def\blend{w}
\def\imagemean{\frac{1}{N}\sum_{(x,y)}}
\def\channelsum{\sum_\textrm{channels}}
\def\disocc{\mathbf{M}}
\newcommand\conv[1]{$\textrm{conv}_{#1}$}
\newcommand\para[1]{\smallskip\noindent\textbf{#1}}
\newcommand\Loss[1]{\mathcal{L}^{\textsf{#1}}}
\newcommand\weight[1]{\lambda_{\textsf{#1}}}
\newcommand\ablation[1]{\textsf{\textbf{\footnotesize#1}}}
\newcommand{\citefig}[1]{Fig.~\ref{#1}}
\newcommand{\citeeq}[1]{Eq.~\ref{#1}}
\newcommand{\citetab}[1]{Table~\ref{#1}}
\newcommand{\citesec}[1]{Section~\ref{#1}}

\begin{abstract}

A recent strand of work in view synthesis uses deep learning to generate multiplane images---a camera-centric, layered 3D representation---given two or more input images at known viewpoints. We apply this representation to \textbf{single-view} view synthesis, a problem which is more challenging but has potentially much wider application. Our method learns to predict a multiplane image directly from a single image input, and we introduce \textbf{scale-invariant view synthesis} for supervision, enabling us to train on online video. We show this approach is applicable to several different datasets, that it additionally generates reasonable depth maps, and that it learns to fill in content behind the edges of foreground objects in background layers.

Project page at \textcolor{blue}{https://single-view-mpi.github.io/}.

\end{abstract}

\section{Introduction}

Taking a photograph and being able to move the camera around is a compelling way to bring photos to life. It requires understanding the 
3D structure of the scene, reasoning about occlusions and what might be behind them, and rendering high quality, spatially consistent new views in real time.

We present a deep learning approach to this task which can be trained on online videos or multi-camera imagery using view synthesis quality as an objective---hence, the approach does not require additional ground truth inputs such as depth. At inference time, our method takes a single RGB image input and produces a representation of a local light field. We adopt the \emph{multiplane image} (MPI) representation, which 
can model disocclusions and non-Lambertian effects, produces views that are inherently spatially consistent, is well-suited to generation by convolutional networks, and can be rendered efficiently in real time~\cite{zhou:2018:stereo}.

Our approach is the first to generate multiplane images directly from a \emph{single} image input, whereas prior work has only estimated MPIs from multiple input views (anywhere from a stereo pair~\cite{zhou:2018:stereo} to twelve images from a camera array~\cite{flynn:2019:deepview}). Compared with multiple-input view synthesis, ours is a much more challenging task. We want the network to learn where different parts of the scene are in space without being able to observe correlations between multiple views, and without any chance to look even a tiny bit `around the corner' of objects. A particular difficulty arises when supervising such a system using view synthesis because of the global scale ambiguity inherent in the input data. We address this with a method of \textit{scale invariant view synthesis} which makes use of sparse point sets produced in the course of generating our training data. We also introduce an \textit{edge-aware smoothness loss} which discourages the depth-maps derived from our predicted MPIs from being unnaturally blurry, even in the absence of depth supervision.

\begin{figure}
\begin{center}
\centering
\vspace{13pt}
\includegraphics[width=\linewidth]{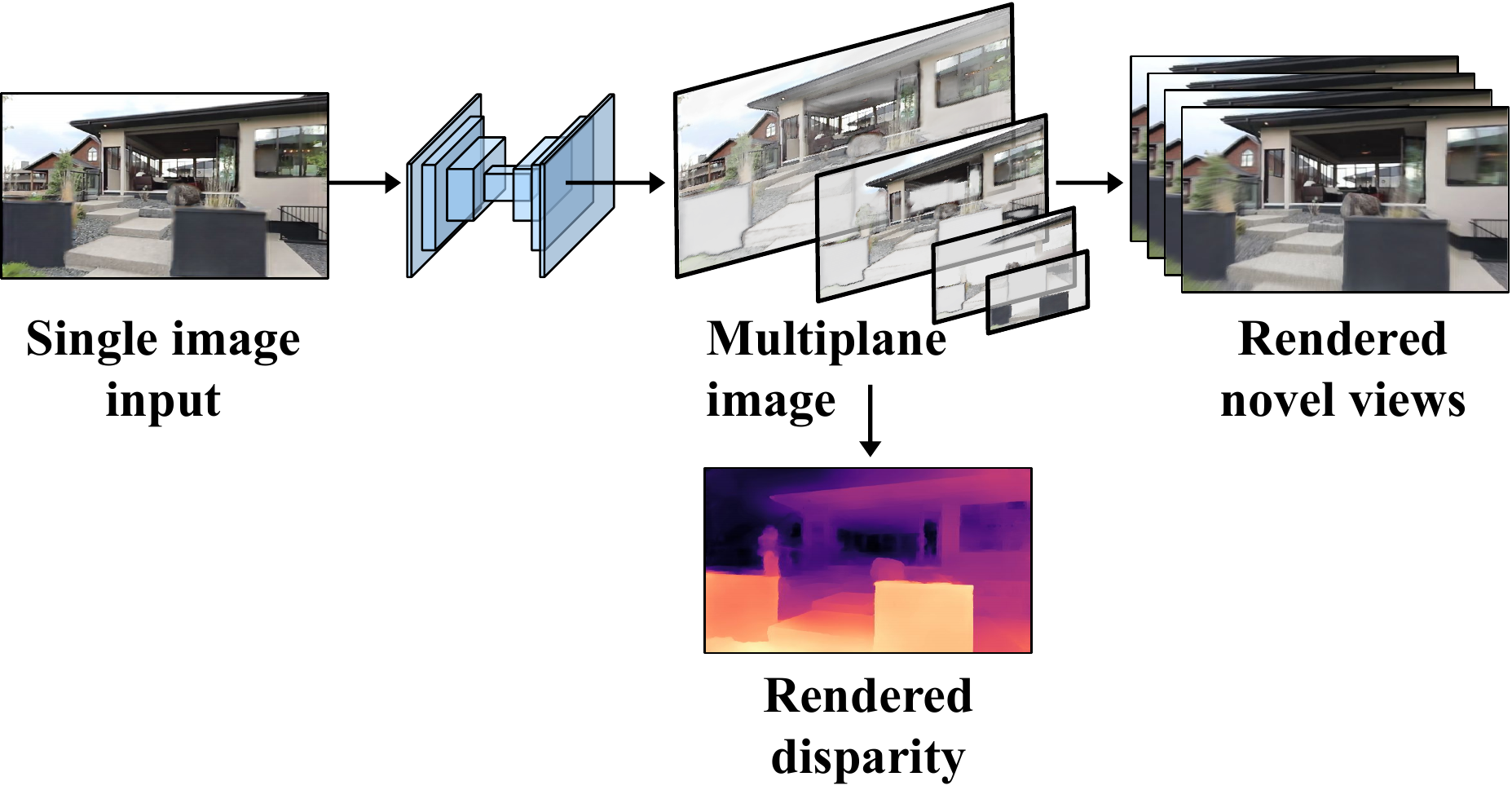}
\caption{Our network generates a multiplane image (MPI) from a single image input. The MPI can be used to render images from novel viewpoints, and to generate a disparity map.  (Video frames here and in other figures are used under Creative Commons license from Youtube user \emph{Sona Visual}.)} 
\label{fig:teaser}
\end{center}
\end{figure}

We train and evaluate our method using a dataset of online videos, and measure the quality of derived depth-estimates on the iBims-1 benchmark. We show the versatility of our approach by comparing it to two previous view synthesis methods using different representations: one that predicts complete 4D light fields and learns from a narrow-baseline multi-angle light field dataset, and another that predicts layered depth images and learns from wide-baseline stereo-camera imagery. Our method not only achieves higher quality view synthesis than these but is more general, not requiring light field data or known-scale inputs for training.

\section{Related work}

We build on work in two areas---view synthesis and depth prediction---that are themselves highly related. Learning-based methods have been applied to both of these domains.

\para{Single-view depth prediction.}
There has been great interest in the task of predicting a depth map, or other suitable geometric representation, from a single RGB image. However, depth maps alone do not enable full synthesis of new views, because they do not capture content that is occluded in the reference view but visible in a desired target view. Conversely, accurate depth is also not strictly required for high-quality view synthesis---for instance, inaccurate depth in textureless regions might be imperceptible, and planarity relationships might be more perceptually important compared to strict accuracy. At the same time, depth and view synthesis are highly intertwined, and many recent depth prediction methods use view synthesis as implicit supervision~\cite{garg:2016:unsupervised,godard:2016:unsupervised,zhou:2017:unsupervised}. Like these methods, we use additional views of scenes as supervision, but we explicitly focus on the application of new view synthesis, and hence use a more expressive scene representation (MPIs) compared to depth maps.

Other recent work, like ours, uses videos in the wild as a source of training data for geometric learning. For instance, Lasinger \etal learn a robust single-view depth predictor from a large dataset of 3D movies, by first extracting optical flow between the left and right frames as a form of pseudo-depth for supervision \cite{lasinger:2019:depth_3d}. Chen \etal produce large quantities of sparse SfM-derived depth measurements from YouTube videos for use in training depth networks~\cite{chen:2019:youtube3d}. However, these prior methods focus on depth, while our method is the first to our knowledge to learn single-view view synthesis from videos in the wild.

\para{Learned view synthesis.}
Traditionally, methods for view synthesis operated in the interpolation regime, where one is provided with multiple views of a scene, and wishes to interpolate views largely within the convex hull of their camera positions. A number of classical approaches to this problem have been explored~\cite{gortler:1996:lumigraph,levoy:1996:lightfield}, including methods that involve estimation of local geometric proxies~\cite{chaurasia:2013:depth,hedman:2018:deepblending,penner:2017:soft3d,zitnick:2004:interpolation}.

Learning-based approaches to this interpolation problem have also been explored. Learning is an attractive tool for view synthesis because a training signal can be obtained simply from having held-out views of scenes from known viewpoints, via predicting those views and comparing to the ground truth images. Some approaches predict new views independently for each output view, leading to inconsistency from one view to the next~\cite{flynn:2016:deepstereo,kalantari:2016:learning}. Other methods predict a single scene representation from which multiple output views can be rendered. In particular, \emph{layered} representations are especially attractive, due to their ability to represent occluded content. For instance, \emph{multiplane images} (MPIs), originally devised for stereo matching problems~\cite{szeliski:1999:stereo_matching_acetates}, have recently found success in both learned view interpolation and extrapolation from multiple input images~\cite{zhou:2018:stereo,srinivasan:2019:boundaries,flynn:2019:deepview,mildenhall:2019:llff}. However, none of these methods are able to predict an MPI from a single input image.

Most related to our work are methods that predict new views from \emph{single} images. This includes work on synthesizing a full light field from a single view~\cite{srinivasan:2017:lightfield}, 
predicting soft disparity maps~\cite{xie:2016:deep3d}, inferring layered representations like layered depth images (LDIs)~\cite{tulsiani:2018:lsi,shade:1998:ldi}, or 
segmenting the input and predicting a 3D plane for each segment~\cite{liu:2018:geometry_aware_single_image_view_synthesis}. We borrow the MPI representation introduced for view inter\-polation and extra\-polation, apply it to the \emph{single-view} case, and show that this representation leads to higher quality results compared to light fields and LDIs.

Depth can also be used as a starting point for view synthesis, as in recent work from Niklaus \etal that predicts a depth map from a single view, then inpaints content behind the visible surfaces to enable high-quality single-image view synthesis~\cite{niklaus:2019:kenburns}. However, this method requires dense, accurate depth supervision and multiple stages of post-processing. Our method learns to predict an MPI as a single stage, using only multiple views (e.g., video frames) as supervision.

\section{Approach}
\label{sec:approach}

At inference time, our method takes a single input image and generates a representation from which novel views at new camera positions can be freely generated (\citefig{fig:teaser}). For training, all we require is videos with static scenes and moving camera, which we process as follows.

\subsection{Data}

We apply SLAM (Simultaneous Localization and Mapping) and structure-from-motion algorithms to videos to identify motion sequences, estimate viewpoints, and generate a sparse point cloud. We follow the method of Zhou \etal \cite{zhou:2018:stereo}: the only difference is that we retain the sparse point cloud and a record of which points were tracked in each frame (referred to as \textit{visible points}), which they do not use.

At training time, we sample pairs of frames (\emph{source} and \emph{target}) from the resulting sequences. Each pair gives us a source image $\Image_s$ and a target image $\Image_t$, together with their viewpoints $v_s$ and $v_t$ (camera intrinsics and extrinsics). Additionally, we extract the set of visible points for the source frame and map them into camera space, resulting in a set $\Points = \{(x, y, d), \ldots\}$ of triples where $(x, y)$ is the position within the source image, and $d$ the depth of that point.

In our experiments, we apply this processing to videos from the RealEstate10K dataset~\cite{zhou:2018:stereo}, giving us over 70000 sequences and over 9 million frames.

\subsection{Representation and rendering}
\label{sec:representation}

\begin{figure}
\begin{center}
\centering
\includegraphics[width=\linewidth]{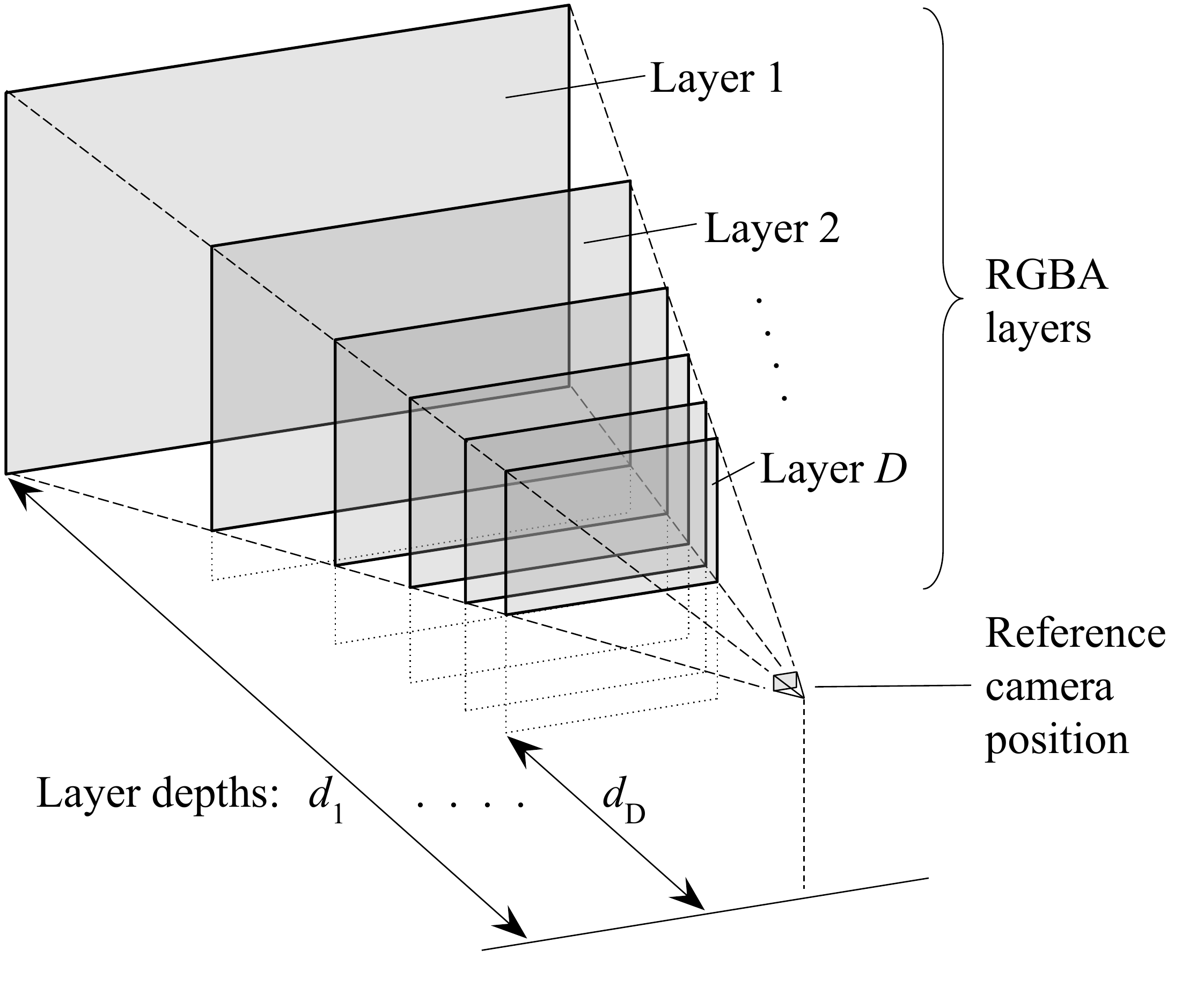}
\caption{The multiplane image representation. See \citesec{sec:representation}.}
\label{fig:mpi}
\end{center}
\end{figure}

We use multiplane images (MPIs) as our scene representation, which support differentiable rendering \cite{zhou:2018:stereo}. As illustrated in \citefig{fig:mpi}, an MPI consists of a set of $D$ fronto-parallel planes in the frustum of a reference camera, arranged at fixed depths $d_1, \ldots, d_D$, from $d_1 = \dfar$ to $d_D = \dnear$, and equally spaced in disparity (inverse depth). Each plane or layer has an RGBA image: we write $c_i$ and $\alpha_i$ for the color and alpha channels of layer $i$, each with a resolution $W \times H = N$. An MPI can also be considered as an instance of the \emph{stack of acetates} model of Szeliski and Golland \cite{szeliski:1999:stereo_matching_acetates} with soft alpha and a specific choice of layer depths.

Given a source image $\Image_s$ at viewpoint $v_s$ our network $\Network$ outputs an MPI whose reference camera is at $v_s$:
\begin{equation}
\{(c_1, \alpha_1), \ldots, (c_D, \alpha_D)\} = \Network(\Image_s).
\end{equation}

\para{Warping.}
The first step in rendering a novel image from an MPI is to warp each layer from the source viewpoint to the desired target viewpoint $v_t$:
\begin{equation}
c'_i = \mathcal{W}_{v_s, v_t}(\scale d_i, c_i),
\hspace{1em}
\alpha'_i = \mathcal{W}_{v_s, v_t}(\scale d_i, \alpha_i).
\label{eq:warp}
\end{equation}
The warping operation $\mathcal{W}$ computes the color or alpha value at each pixel in its output by sampling bilinearly from the input color or alpha. To do this, it applies a homography to each target pixel's coordinates $(u_t, v_t)$ to obtain corresponding source coordinates $(u_s, v_s)$ at which to sample:
\begin{equation}
\begin{bmatrix} u_s \\ v_s \\ 1 \end{bmatrix}
~ \sim ~
~ K_s~ \left(R - \frac{\mathbf{t}\mathbf{n}^\mathsf{T}}{\mathbf{a}}\right) ~K_t^{-1}
\begin{bmatrix} u_t \\ v_t \\ 1 \end{bmatrix},
\label{eq:homography}
\end{equation}
where $\mathbf{n}$ is the normal vector and $\mathbf{a}$ the distance (both relative to the target camera) to a plane that is fronto-parallel to the source camera at depth $\scale d_i$, $R$ is the rotation and $\mathbf{t}$ the translation from $v_t$ to $v_s$, and $K_s$, $K_t$ are the source and target camera intrinsics.

This procedure and the compositing that follows are the same as in Zhou \etal \cite{zhou:2018:stereo}, except for the introduction of the scale factor $\scale$ in \citeeq{eq:warp} and \citeeq{eq:homography}. The layer depths $d_i$ are multiplied by $\scale$, scaling the whole MPI up or down correspondingly. As described in \citesec{sec:scale}, choosing the right scale $\scale$ allows us to overcome the scale ambiguity inherent to SfM models and achieve scale-invariant synthesis.

\para{Compositing.}
The warped layers $(c'_i, \alpha'_i)$ are composited using the \textit{over} operation \cite{porter:1984:compositing_over} to give the rendered image $\Imagehat_t$:
\begin{equation}
  \Imagehat_t = \sum_{i = 1}^{D} \Bigl(c'_i \alpha'_i \prod_{j=i+1}^{D} (1 - \alpha'_j)\Bigr).
\label{eq:over}
\end{equation}
We can also synthesize a disparity-map $\Disparity_s$ from an MPI, by compositing the layer disparities (i.e.~inverse depths):
\begin{equation}
  \Disparity_s = \sum_{i = 1}^{D} \Bigl(d_i^{-1} \alpha_i \prod_{j=i+1}^{D} (1 - \alpha_j)\Bigr).
\label{eq:disparity}
\end{equation}
Note that although layer depths are discrete, the disparity-map can be smooth because $\alpha_i$ blends softly between layers.

\begin{figure}
\begin{center}
\centering
\includegraphics[width=\linewidth]{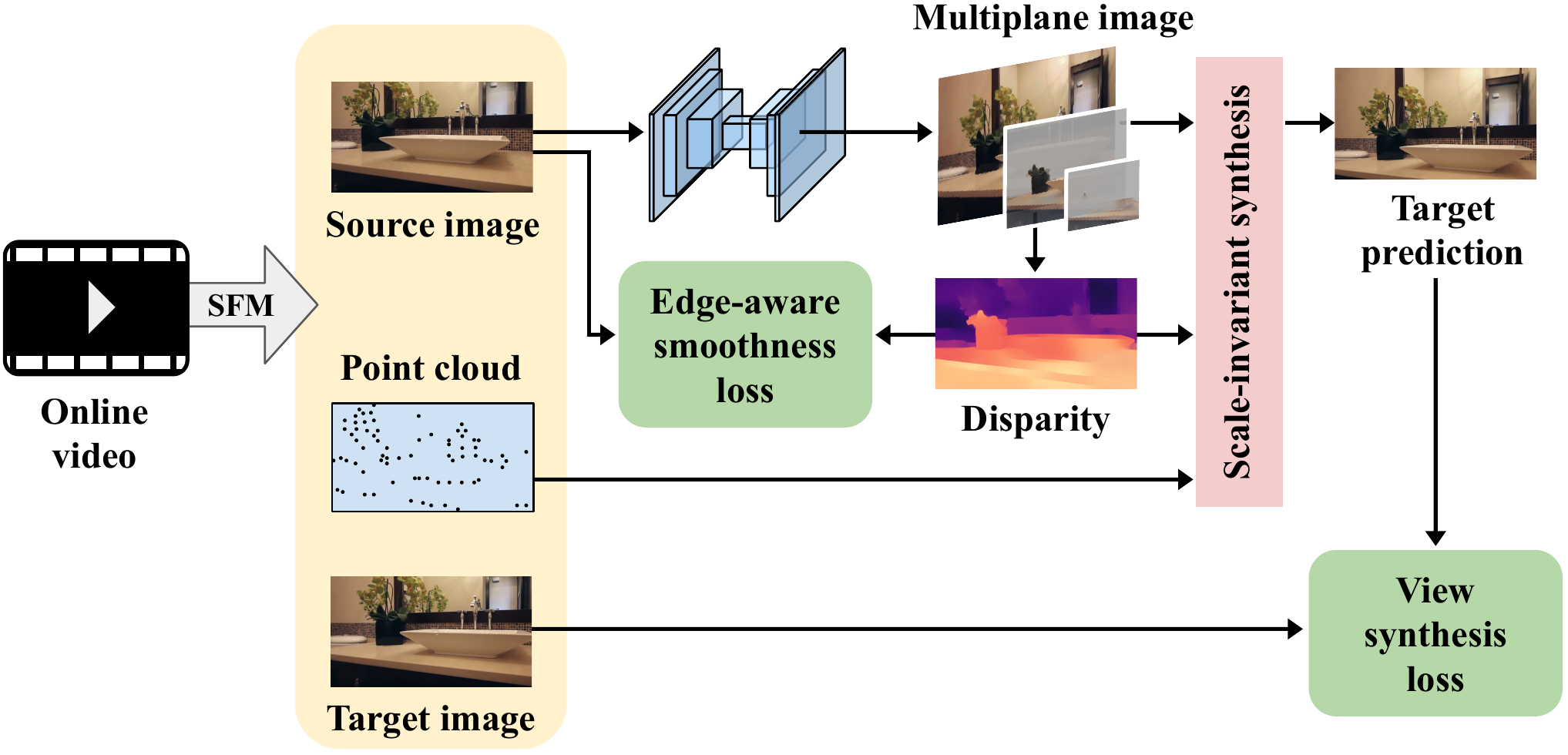}
\caption{Our system, trained on online video, learns to predict multiplane images directly from single image inputs. Scale-invariant view synthesis allows us to apply view synthesis loss despite the global scale ambiguity in our training data.} 
\label{fig:system}
\end{center}
\end{figure}

\subsection{Scale-invariant synthesis}
\label{sec:scale}
Visual SLAM and structure-from-motion have no way of determining absolute scale without external information: each of our training sequences is therefore equally valid if we scale the world (including the sparse point sets and the translation part of the camera poses) up or down by any constant factor. This is not an issue when dealing with multiple-image input since the relative pose between the inputs resolves the scale ambiguity, but it poses a challenge for learning any sort of 3D representation from a \emph{single} input. To address this ambiguity, prior work on single-view depth prediction typically employs a \emph{scale-invariant depth loss} \cite{eigen:2014:depth_map_prediction,ummenhofer:2017:demon} or more recently even a \emph{scale-and-shift invariant depth loss} \cite{lasinger:2019:depth_3d}. These methods can be seen as finding the scale factor which minimizes a scale-dependent loss, and rely on there being a closed-form solution for this scale factor.

View-synthesis losses are subject to the same problem: the scale of each training instance is arbitrary, but without the correct scale, rendered images cannot match ground truth. We could try to minimize a view synthesis loss over all possible scale factors, but the rendering operations (Eqs.~\ref{eq:warp}--\ref{eq:over}) make this not amenable to a closed-form solution, and direct optimization for scale at run-time will be prohibitively slow.

We observe that although scale is unknown, the camera poses $v_s$, $v_j$ and the point set $\Points_s$ do have a \textit{consistent} scale for each training example. Therefore we can use the point set to compute a scale factor to apply in rendering. We compute the scale factor $\scale$ which minimizes log-squared error between the predicted disparity $\Disparity_s$ and the point set $\Points_s$:
\begin{equation}
\scale = \exp\left[\frac{1}{|\Points_s|}\sum_{(x,y,d) \in \Points_s}(\ln\Disparity_s(x, y) - \ln (d^{-1}))\right]
\label{eq:bestscale}
\end{equation}
where $\Disparity_s(x,y)$ denotes bilinear sampling from the disaparity map at position $(x,y)$. The scale factor $\scale$ thus obtained is applied in Eqs.~\ref{eq:warp} and \ref{eq:homography}, ensuring that the rendered image $\Imagehat_t$ no longer varies with the scale of the input viewpoints and point set. $\Imagehat_t$ is therefore suitable for use in training with view-synthesis losses.

\subsection{Losses}

Our overall loss combines a view synthesis loss, a smoothness loss on the synthesized disparity, and a sparse depth supervision loss:
\begin{equation}
\Loss{} = \weight{p}\Loss{pixel}
  + \weight{s}\Loss{smooth} + \weight{d}\Loss{depth}
\end{equation}
We now describe each of these in turn.

\para{Synthesis.}
To encourage the rendered image at the target viewpoint to match the ground truth, we use an $L^1$ per-pixel loss:
\begin{align}
\label{eq:synthesisloss}
\Loss{pixel} &= \channelsum \imagemean |\Imagehat_t - \Image_t|.
\end{align}
We can optionally add an image gradient term to this, but we did not find it to be consistently helpful.

\para{Edge-aware smoothness.}
For natural images, depth discontinuities are typically accompanied by discontinuities in the image itself (though the reverse is not the case) \cite{gamble:1987:visual_integration_discontinuities}. This idea has been used in classical computer vision, notably in stereo correspondence \cite{scharstein:2002:taxonomy_correspondence}, and also in a variety of different smoothness losses for learning depth prediction \cite{godard:2016:unsupervised,li:2019:moving_people,wang:2018:direct_methods}.
These losses work by encouraging depth to be smooth wherever the input image is smooth.

We apply this idea as follows. First, let $\Sobel$ be the sum over all channels of the $L^1$ norm of the gradient of an image (we use Sobel filters to compute the gradient):
\begin{equation}
    \Sobel(\Image) = \channelsum \big\|\nabla \Image\big\|_1 
\end{equation}
We define a \textit{source edge mask} $\mathbf{E}_s$ which is 1 wherever the source image gradient is at least a fraction $\emin$ of its maximum over the image.
\begin{equation}
    \mathbf{E}_s = \min\Bigl(\frac{\Sobel(\Image_s)}{\emin \times \max_{(x,y)}\Sobel(\Image_s)}, 1\Bigr)
\end{equation}
Our \textit{edge-aware smoothness loss} then penalizes gradients higher than a threshold $\gmin$ in the predicted disparity map, but only in places where the edge mask is less than one:
\begin{equation}
\Loss{smooth} = \imagemean \left(\max(\Sobel(\Disparity_s) - \gmin, 0) \odot (1 - \mathbf{E}_s)\right),
\end{equation}
where $\odot$ is the Hadamard product. As with our synthesis loss, $\Loss{smooth}$ is an average over all pixels. In practice we set $\emin = 0.1$ and $\gmin = 0.05$.

As noted earlier, there are many possible formulations of such a loss. Our $\Loss{smooth}$ is one that we found creates qualitatively better depth maps in our system, by allowing gradual changes in disparity while encouraging discontinuities to be accurately aligned to image edges.

\para{Sparse depth supervision.}
The point set $\Points_s$ allows us to apply a form of direct but sparse depth supervision. We adopt the $L^2$ loss of Eigen \etal on log disparity \cite{eigen:2014:depth_map_prediction} (as noted in \citesec{sec:scale}, $\sigma$ is the scale factor that minimizes this loss---it is equivalent to the variable $\alpha$ in Eigen \etal's scale-invariant loss, under $\ln \scale = \alpha$):
\begin{equation}
\Loss{depth} = \frac{1}{|\Points_s|}\sum_{(x,y,d) \in \Points_s}\Bigl(\ln\frac{\Disparity_s(x, y)}{\scale} - \ln (d^{-1})\Bigr)^2
\end{equation}

\subsection{Implementation}
\label{sec:implementation}

\begin{table}
\centering
\begin{tabular}{@{}l|rrrrl@{}}
\toprule
Input & $\mathrm{k}_1$ & $\mathrm{c}_1$ & $\mathrm{k}_2$ & $\mathrm{c}_2$ & Output
\tabularnewline
\midrule
$\Image_s$                   & 7 &  32 & 7 &  32 & \conv{1}   \tabularnewline
  \down(\conv{1})            & 5 &  64 & 5 &  64 & \conv{2}   \tabularnewline
  \down(\conv{2})            & 3 & 128 & 3 & 118 & \conv{3}   \tabularnewline
  \down(\conv{3})            & 3 & 256 & 3 & 256 & \conv{4}   \tabularnewline
  \down(\conv{4})            & 3 & 512 & 3 & 512 & \conv{5}   \tabularnewline
  \down(\conv{5})            & 3 & 512 & 3 & 512 & \conv{6}   \tabularnewline
  \down(\conv{6})            & 3 & 512 & 3 & 512 & \conv{7}   \tabularnewline
  \down(\conv{7})            & 3 & 512 & 3 & 512 & \conv{8}   \tabularnewline

  \up(\conv{8}) + \conv{7}   & 3 & 512 & 3 & 512 & \conv{9}   \tabularnewline
  \up(\conv{9}) + \conv{6}   & 3 & 512 & 3 & 512 & \conv{10}  \tabularnewline
  \up(\conv{10}) + \conv{5}  & 3 & 512 & 3 & 512 & \conv{11}  \tabularnewline
  \up(\conv{11}) + \conv{4}  & 3 & 512 & 3 & 512 & \conv{12}  \tabularnewline
  \up(\conv{12}) + \conv{3}  & 3 & 128 & 3 & 128 & \conv{13}  \tabularnewline
  \up(\conv{13}) + \conv{2}  & 3 &  64 & 3 &  64 & \conv{14}  \tabularnewline
  \up(\conv{14}) + \conv{1}  & 3 &  64 & 3 &  64 & \conv{15}  \tabularnewline
  \conv{15}                  & 3 &  64 & 3 &  64 & \conv{16}  \tabularnewline

  \conv{16}                  & 3 &  34 & - & - & output  \tabularnewline
\bottomrule
\end{tabular}
\vspace{10pt}

\caption{Our network architecture. Each row describes \emph{two} convolutional layers in sequence: $\mathbf{k}_1, \mathbf{k}_2$ are the kernel sizes and $\mathbf{c}_1, \mathbf{c}_2$ the numbers of output channels. \textbf{Input} shows the input to the first layer, where \down\ denotes maxpooling with a pool size of 2 (thus halving the size), \up\ denotes nearest-neighbour upscaling by a factor of 2, and $+$ is concatenation. Each layer is followed by ReLU activation. The final row shows a single convolutional layer, which is instead followed by sigmoid activation. For details of how the outputs are translated into MPI layers, see \citesec{sec:implementation}.}

\label{tab:network}
\end{table}

\begin{table*}
\centering
\begin{tabular}{@{}lcccccccccccccc@{}}
\toprule
& \multicolumn{2}{c}{LPIPS$_\textrm{all}\downarrow$}
&& \multicolumn{2}{c}{PSNR$_\textrm{all}\uparrow$} && \multicolumn{2}{c}{SSIM$_\textrm{all}\uparrow$}
&& \multicolumn{2}{c}{PSNR$_\textrm{disocc}\uparrow$} && \multicolumn{2}{c}{SSIM$_\textrm{disocc}\uparrow$} \\
\cmidrule{2-3} \cmidrule{5-6} \cmidrule{8-9} \cmidrule{11-12} \cmidrule{14-15}
Method &  $n=5$ & $n=10$ && $n=5$ & $n=10$ && $n=5$ & $n=10$ && $n=5$ & $n=10$ && $n=5$ & $n=10$ \tabularnewline
\midrule
\ablation{full}          & 0.103 & \textbf{0.155} && 26.4 & 23.5  &&  0.859 & 0.795  &&  \textbf{19.7} & 17.9  &&  0.513 & 0.480  \tabularnewline
\ablation{nodepth}       & 0.120 & 0.178 && 26.2 & 23.4  &&  0.854 & 0.791  &&  19.2 & 18.0  &&  0.525 & 0.496  \tabularnewline
\ablation{noscale}       & 0.149 & 0.221 && 25.4 & 22.8  &&  0.837 & 0.771  &&  18.5 & 17.3  &&  0.496 & 0.470  \tabularnewline
\ablation{nosmooth}      & 0.104 & 0.159 &&  26.4 & 23.6  &&  0.860 & 0.798  &&  19.6 & \textbf{18.4}  &&  \textbf{0.540} & \textbf{0.527}  \tabularnewline
\ablation{nobackground}  & \textbf{0.099} & 0.162 && \textbf{26.8} & \textbf{23.7}  &&  \textbf{0.867} & \textbf{0.802}
                         &&  18.7 & 17.7  &&  0.509 & 0.499  \tabularnewline
\bottomrule
\end{tabular}
\vspace{10pt}
\caption{Ablation studies on images from RealEstate10K video sequences. $n$ indicates the number of frames between source and target in the video sequence. `\emph{all}' metrics are computed on the whole image (with a 5\% crop), `\emph{disocc}' metrics on disoccluded pixels only, i.e.~those where $\disocc_t > 0.6$. We observe that scale-invariance gives a large benefit, depth supervision a smaller one, and that predicting background content does not clearly help overall but does improve performance for disoccluded pixels and on perceptual similarity. See~\citesec{sec:ablations}.}

\label{tab:ablations}
\end{table*}

\noindent \textbf{Network.}
We use a DispNet-style network \cite{mayer:2016:dispnet}, specified in \citetab{tab:network}. We pad the input (the single RGB image $\Image_s$) to a multiple of 128 in height and width, and crop the output correspondingly. The first $D - 1$ channels of the output give us $\alpha_2, \ldots, \alpha_D$. The back layer is always opaque, so $\alpha_1 = 1$ and need not be output from the network. When initializing our network for training, we set the bias weights on the last convolutional layer so that the mean of the initial output distribution corresponds to an initial alpha value of $1/i$ in layer $i$. This \emph{harmonic bias} helps ameliorate an issue during training in which layers which are not near the front of the MPI volume are heavily occluded and have very small gradients with respect to our losses.

We follow Zhou~\etal\cite{zhou:2018:stereo} and model each layer's color as a per-pixel blend of the input image with a predicted global \emph{background image} $\Imagehat_\textit{bg}$. In that work, blend weights are predicted for each pixel in each MPI layer. We reason instead that content that is visible (from the source viewpoint) should use the foreground image, and content that is fully occluded should use the background image. Therefore we can derive the blend weights $\blend_i$ from the alpha channels as follows:
\begin{align}
\blend_i &= \prod_{j > i} (1 - \alpha_j), \\
c_i &= \blend_i \Image_s + (1 - \blend_i) \Imagehat_{\textit{bg}}.
\end{align}
The background image is determined by the remaining three channels of the network output. Because it is difficult for the network to learn to predict $\alpha_i$ and $\Imagehat_{\textit{bg}}$ simultaneously, during training we set $\Imagehat_\textit{bg}$ to be a linear interpolation between $\Image_s$ and the network output, with the contribution of the network increasing gradually over the first $\blendsteps$ training steps.

\para{Training.}
In our experiments, $D$ (the number of MPI planes) is $32$, $\blendsteps = 100,\!000$, and our losses are weighted as follows: $\weight{p} = 1$, $\weight{s} = 0.5$, $\weight{p} = 0.1$. We train using the Adam Optimizer \cite{kingma:2014:adam} with a learning rate of 0.0001.

\section{Experiments}

We present quantitative and qualitative evaluations of our method on the RealEstate10K dataset, depth evaluations with the iBims-1 benchmark, and comparisons with previous view synthesis methods on the Flowers and KITTI datasets. Because of the very visual nature of the view synthesis task, we strongly encourage the reader to view the additional examples, including animations, in our supplementary video.

\subsection{View synthesis on RealEstate10K}
\label{sec:ablations}

To investigate the effects of our different losses and MPI background prediction, we train several versions of our method on videos from the RealEstate10K dataset \cite{zhou:2018:stereo}:
\begin{itemize}[noitemsep]
\item\ablation{full}:
Our full method, as described in \citesec{sec:approach}.
\item\ablation{nodepth}:
As \ablation{full}, but with no depth loss, i.e.~$\weight{d} = 0$.
\item\ablation{noscale}:
As \ablation{full}, but with no depth loss and no scale-invariance, i.e.~$\weight{d} = 0$, $\scale = 1$.
\item\ablation{nosmooth}:
As \ablation{full}, but with no edge-aware disparity smoothness loss, i.e.~$\weight{s} = 0$.
\item\ablation{nobackground}:
As \ablation{full}, but with no background prediction. Instead, all MPI layers take their color from the input, i.e.~$c_i = \Image_s$.
\end{itemize}
To compare these methods, we measure the accuracy of synthesized images using the LPIPS perceptual similarity metric \cite{zhang:2018:lpips} and PSNR and SSIM metrics, on a held-out set of 300 test sequences, choosing source and target frames to be 5 or 10 frames apart. At test time, we use the point set to compute the scale factor $\scale$ in the same way as we do during training---for a fair comparison, we also do this for the \ablation{noscale} model. Results are in \citetab{tab:ablations} (LPIPS$_\textrm{all}$, PSNR$_\textrm{all}$ and SSIM$_\textrm{all}$ columns).

We observe that \ablation{nodepth} performs a little worse than \ablation{full}, and \ablation{noscale} performs considerably worse still. This shows that direct depth supervision---although sparse---is of some benefit, but that the improvement from our \emph{scale-invariant synthesis} is more significant. As expected, for all variants, performance decreases with larger camera movement.

Somewhat unintuitively, the \ablation{nosmooth} and \ablation{nobackground} models outperform the \ablation{full} model on PSNR and SSIM metrics. But at larger distances the LPIPS metric, which attempts to measure perceptual similarity, shows benefits from our smoothness loss and from allowing the network to predict the background layer, with the \ablation{full} model performing best.

Qualitatively, the \ablation{nobackground} model introduces unpleasant artefacts at the edges of foreground objects, whereas the \ablation{full} model is able to use the background layer to predict the appearance of some disoccluded content, as shown in \citefig{fig:background}.
To quantify this effect, we first compute a \emph{disocclusion mask} $\disocc_t$ for each image by warping and compositing the blend weights $\blend_i$ used in the \ablation{full} model:

\begin{figure}[t]
\begin{center}
\includegraphics[width=1.0\linewidth]{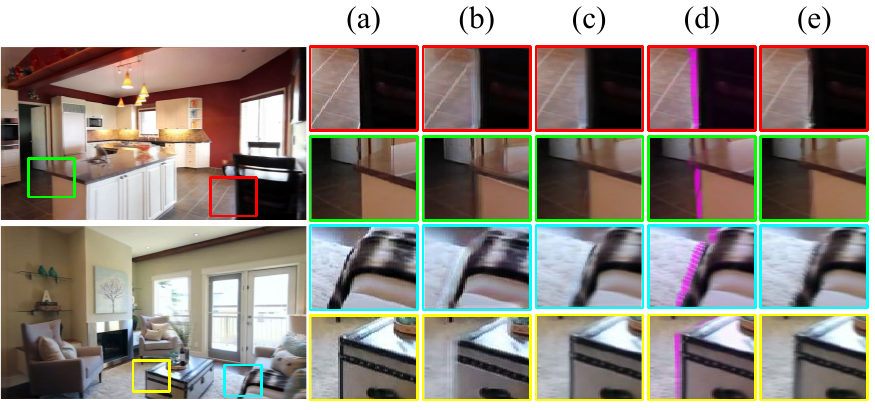}
\end{center}
\caption{Use of the background image. For each region we show (a) the input image $\Image_s$, (b) the predicted background $\Imagehat_{bg}$, (c) a rendering generated by our \ablation{full} method from a slightly different viewpoint, (d) a visualization of where $\Imagehat_{bg}$ is used in rendering: pixels whose value comes 90\% or more from the background are highlighted, (e) the same region rendered by our \ablation{nobackground} model. Comparing (c) and (e), images rendered by our \ablation{full} model show cleaner edges with fewer artefacts than those rendered by the \ablation{nobackground} model. Comparing (a) and (b), the network has learned to erode the edges of foreground objects and to predict what color may be behind them, although some artefacts remain.}
\vspace{-20pt}
\label{fig:background}
\end{figure}
\begin{align}
\blend'_i &= \mathcal{W}_{v_s, v_t}(\scale d_i, \blend_i), \nonumber \\
\disocc_t &= 1 - \sum_{i = 1}^{D} \Bigl(\blend'_i \alpha'_i \prod_{j=i+1}^{D} (1 - \alpha'_j)\Bigr).
\end{align}
$\disocc_t$ tells us how much of the composited value at each pixel comes from the background image. We use it to compute metrics only on diosccluded pixels, i.e.~those where $\disocc_t$ is greater than some threshold. Results are in \citetab{tab:ablations} (PSNR$_\textrm{disocc}$ and SSIM$_\textrm{disocc}$ columns). Although \ablation{nobackground} achieves slightly better scores than \ablation{full} over the whole image, it performs worse on these disoccluded areas.

The \ablation{nosmooth} model achieves plausible view synthesis results, but this is not the only potential application of MPIs. For other tasks, such as editing or object insertion, it is desirable to have accurate depth maps. As shown in \citefig{fig:smoothness}, \ablation{nosmooth} performs significantly worse than our full model in this regard: it both lacks sharp edges where depth should be discontinuous, and introduces discontinuities where depth should be smooth. Our point set depth data is insufficient to evaluate depth accuracy on the RealEstate10K dataset, so for a quantitative measurement we turn to another benchmark.

\begin{figure}[t!]
\begin{center}
\includegraphics[width=1.0\linewidth]{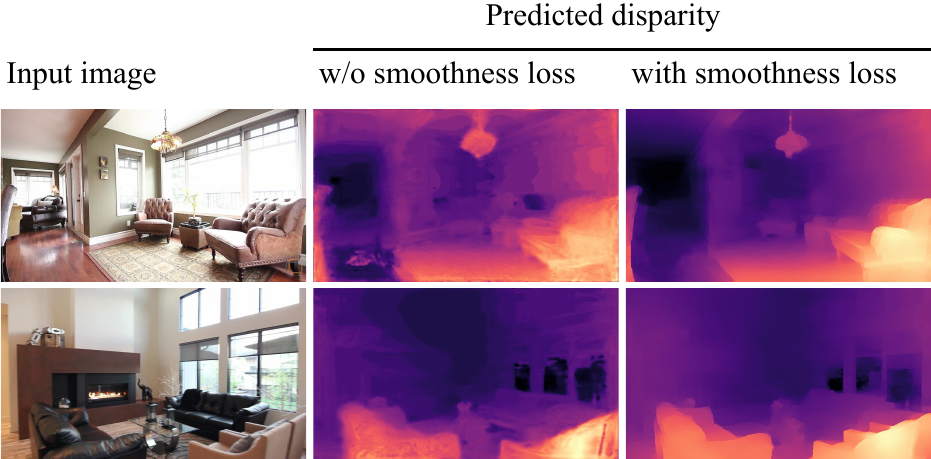}
\end{center}
\caption{Effect of smoothness loss on predicted disparity. As shown in these examples, our edge-aware smoothness loss encourages the predicted disparity to be smooth where the input image is smooth, and consequently also encourages it to have sharp edges aligned with visible object boundaries.}
\label{fig:smoothness}
\end{figure}

\subsection{Depth evaluation}
\label{sec:ibims}
\begin{table}
\centering
\resizebox{\linewidth}{!}{\begin{tabular}{@{}lrrrrrr@{}}
\toprule
Method & rel$\,\downarrow$ & log10$\,\downarrow$ & RMS$\,\downarrow$ &
$\sigma_1\uparrow$ & $\sigma_2\uparrow$ & $\sigma_3\uparrow$

\tabularnewline
\midrule
DIW &
   0.25 & 0.10 & 1.00 & 0.61 & 0.86 & 0.95   \tabularnewline 
MegaDepth (Mega) &
   0.23 & 0.09 & 0.83 & 0.67 & 0.89 & 0.96   \tabularnewline
MegaDepth (Mega + DIW) &
   0.20 & 0.08 & 0.78 & 0.70 & 0.91 & 0.97   \tabularnewline
\vspace{4pt}
3DKenBurns &
   \textbf{0.10} & \textbf{0.04} & \textbf{0.47} & \textbf{0.90} & \textbf{0.97} & \textbf{0.99}   \tabularnewline
Ours: \ablation{full} &
   0.21 & 0.08 & 0.85 & 0.70 & 0.91 & 0.97   \tabularnewline
\hspace{.5em}\ablation{nodepth} ($\weight{depth} = 0$) &
   0.23 & 0.09 & 0.90 & 0.67 & 0.89 & 0.96   \tabularnewline
\hspace{.5em}\ablation{noscale} ($\scale = 1$) &
   0.23 & 0.09 & 0.89 & 0.65 & 0.89 & 0.97   \tabularnewline
\hspace{.5em}\ablation{nosmooth} ($\weight{smooth} = 0)$ &
   0.24 & 0.09 & 0.94 & 0.65 & 0.87 & 0.96   \tabularnewline
\hspace{.5em}\ablation{nobackground} ($c_i = \Image_s$) &
   0.22 & 0.09 & 0.90 & 0.67 & 0.90 & 0.97   \tabularnewline
\bottomrule
\end{tabular}}
\vspace{10pt}
\caption{Measuring depth prediction quality with the iBims-1 benchmark \cite{koch:2018:ibims}.
While not state of the art in terms of depth prediction, our method is comparable to other systems that use explicit depth supervision, even when we use no depth supervision at all. We reran the MegaDepth model to ensure consistency; results for other methods are as reported by Niklaus \etal~\cite{niklaus:2019:kenburns}. See~\citesec{sec:ibims}.}

\label{tab:ibims}
\end{table}

While our objective is view synthesis not depth prediction, we can conveniently synthesize disparity maps from our MPIs, and use them to evaluate depth performance. Here we measure this using the iBims-1 benchmark \cite{koch:2018:ibims}, which has ground truth depth for a variety of indoor scenes. As in Niklaus \etal, we scale and bias depth predictions to minimize the ($L^2$) depth error before evaluation  \cite{niklaus:2019:kenburns}. In \citetab{tab:ibims}, we compare performance with three depth-prediction methods: MegaDepth \cite{li:2018:megadepth}, Depth in the Wild \cite{swirski:2011:layered_photo_popup} and the recent ``3D Ken Burns Effect'' system \cite{niklaus:2019:kenburns}. Of our models, the \ablation{full} version performs the best, at a level comparable to MegaDepth, despite that method's much heavier reliance on explicit depth supervision. As we would expect, removing depth supervision and/or scale invariance leads to worse performance. Our \ablation{nosmooth} model performs the worst, confirming that our edge-aware smoothness loss is valuable in learning to predict MPIs that correspond to good depth-maps.

\begin{figure}[t]
\centering
\includegraphics[width=1.0\linewidth]{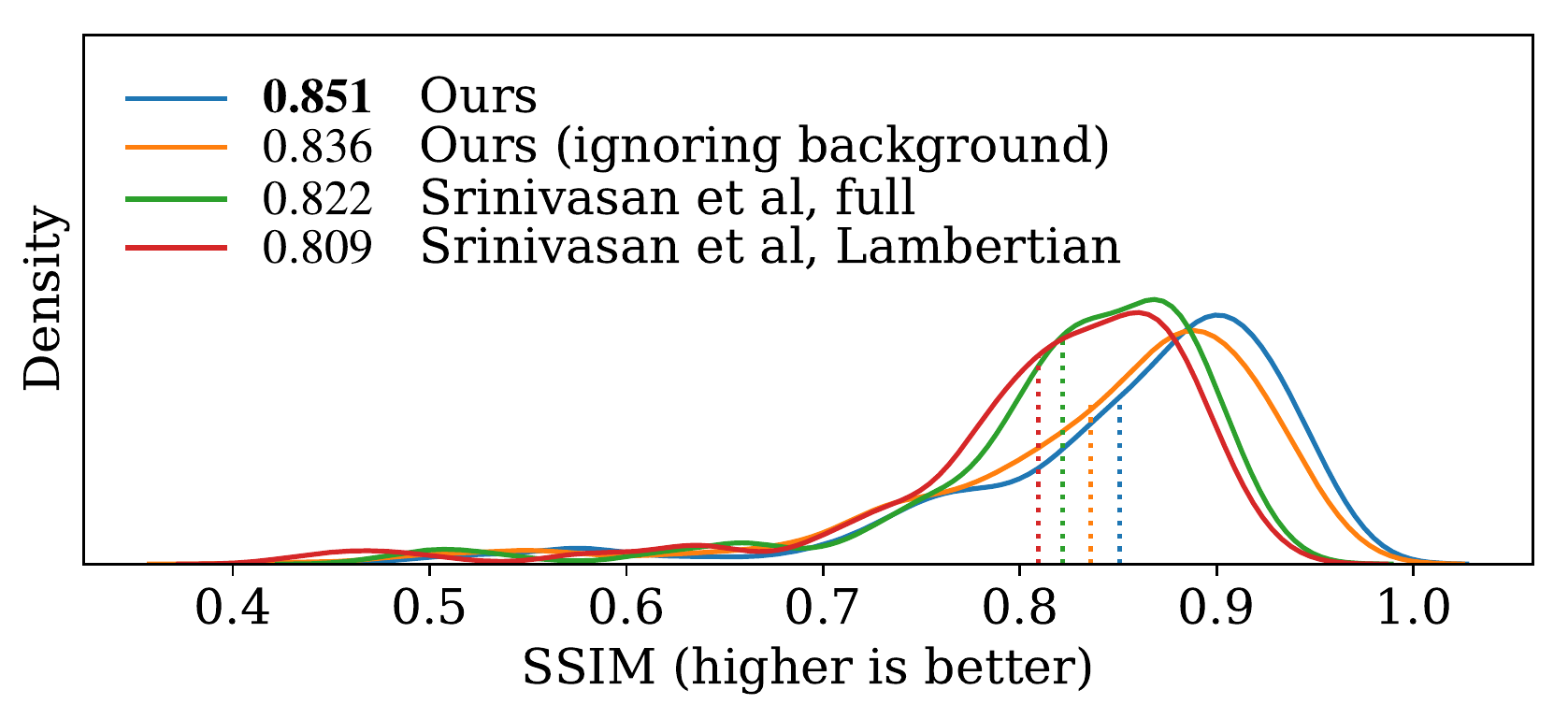}
\includegraphics[width=1.0\linewidth,trim={0 0 0 1em}]{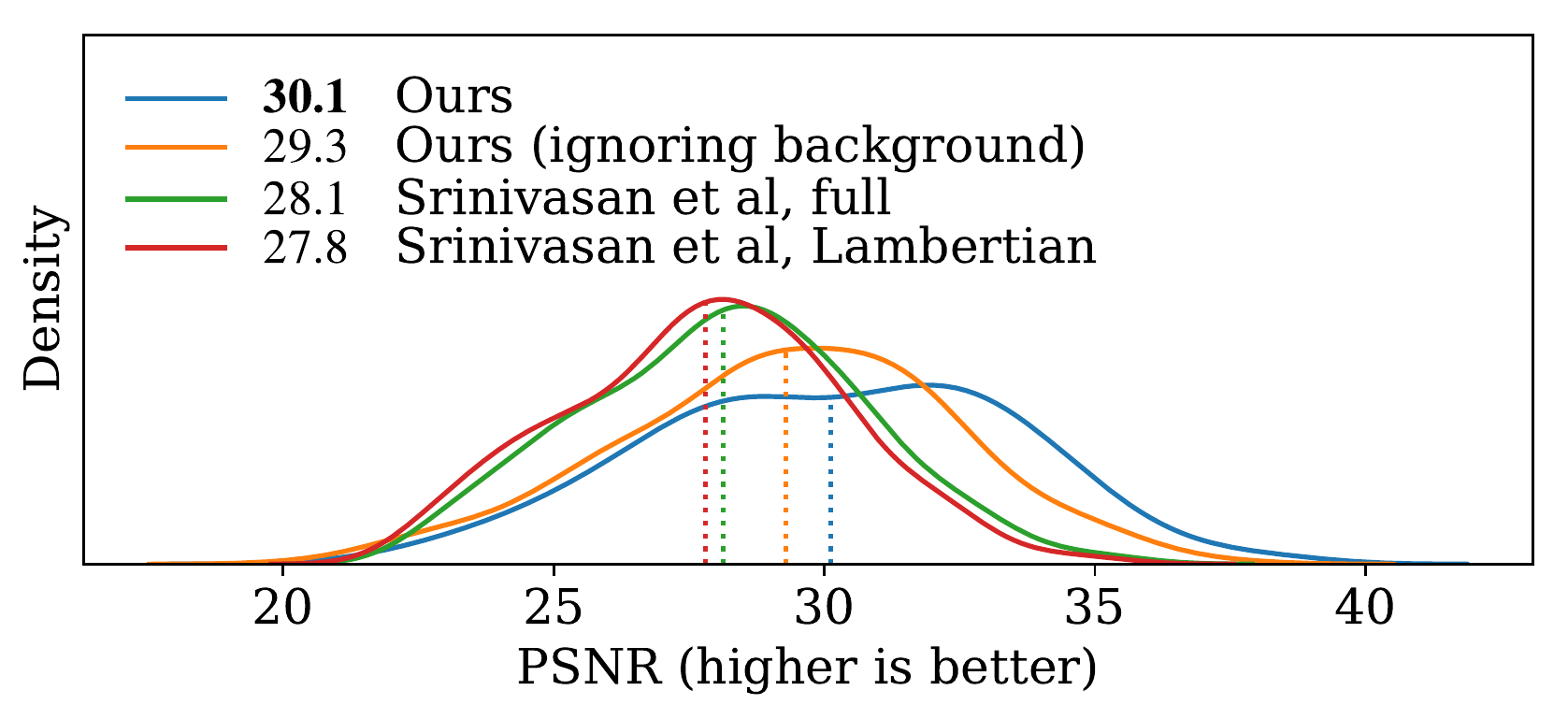}
\caption{Results of running our method on the Flowers dataset, inputting a central view and synthesizing views from the four corner angles. We visualise the distribution of SSIM and PSNR metrics across the test set of 100 light fields (dotted lines and numbers in the legend show the mean values). See~\citesec{sec:flowers}.}
\label{fig:flowers}
\end{figure}

\begin{figure}[t]
\centering
\includegraphics[width=1.0\linewidth,trim={0 0 0 1em}]{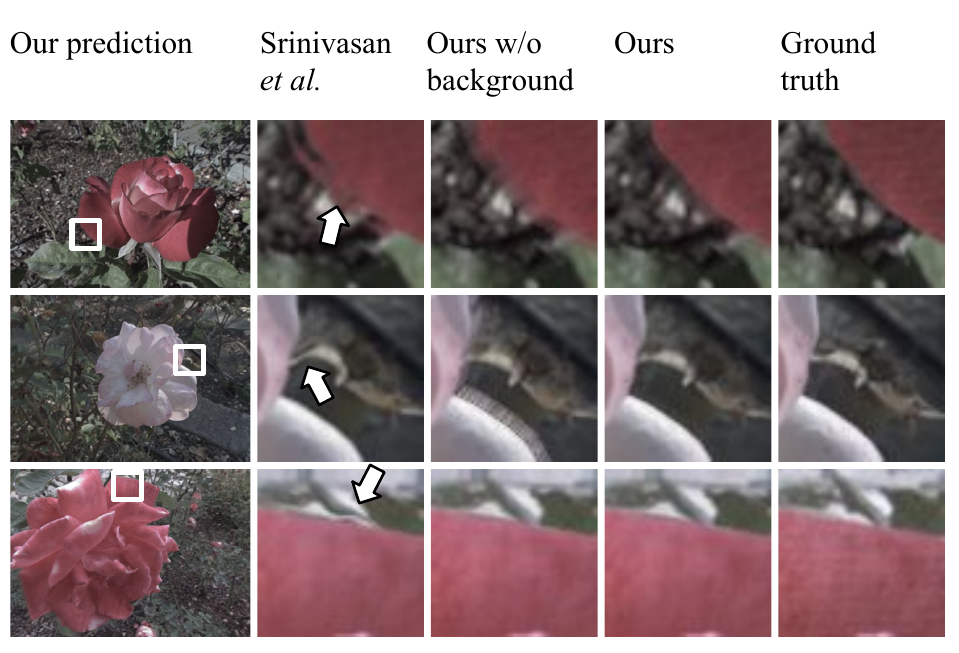}
\caption{Comparisons on Flowers light fields. Views rendered by our method without using the predicted background show blurriness and repeated edge artefacts at depth boundaries, which the predicted background ameliorates. Our predictions improve on those of Srinivasan \etal by avoiding occasional large `floating' foreground artefacts, and by reducing the distortion of background elements. See~\citesec{sec:flowers}.}
\label{fig:flowers_examples}
\end{figure}

\begin{table}[t]
\begin{threeparttable}
\small
\centering
\begin{tabularx}{\linewidth}{@{}Xccccc@{}}
\toprule
 & \multicolumn{2}{c}{PSNR$\;\uparrow$} &&  \multicolumn{2}{c}{SSIM$\;\uparrow$} \tabularnewline
\cmidrule{2-3} \cmidrule{5-6}
Method & all & disocc. && all & disocc. \tabularnewline
\midrule
\addlinespace[4pt]
\multicolumn{4}{l}{\emph{Evaluated at} 384 $\times$ 128 \emph{pixels}} \tabularnewline
\addlinespace[2pt]
Tulsiani~\etal\tnote{1}      & 16.5 & 15.0 && 0.572 & 0.523 \tabularnewline
Ours (`low res')\tnote{2}    & 19.3 & 17.2 && 0.723 & 0.631 \tabularnewline
\textbf{Ours (full)}\tnote{3}  & \textbf{19.5} & \textbf{17.5} && \textbf{0.733} & \textbf{0.639} \tabularnewline
\hspace*{1em}w/o background & 19.3 & 16.9 && 0.731 & 0.627 \tabularnewline

\midrule
\addlinespace[4pt]
\multicolumn{4}{l}{\emph{Evaluated at} 1240 $\times$ 375 \emph{pixels}} \tabularnewline
\addlinespace[2pt]
\textbf{Ours (full)} & \textbf{19.3} & \textbf{17.4} && \textbf{0.696} & \textbf{0.651} \tabularnewline
\hspace*{1em}w/o background & 19.1 & 16.7 && 0.690 & 0.634 \tabularnewline

\bottomrule
\end{tabularx}

\begin{tablenotes}\small{
\item[1] Their method predicts layers at 768 $\times$ 256 resolution but renders at 384 $\times$ 128 to avoid cracks.
\item[2] For a fair comparison, our `low res' model is trained to predict layers at 768 $\times$ 256.
\item[3] Our full method predicts layers at 1240 $\times$ 375.}
\end{tablenotes}
\vspace{10pt}

\end{threeparttable}
\caption{Evaluation on city sequences from the KITTI dataset. We compute PSNR and SSIM metrics over all pixels and over `disoccluded' pixels only. The upper part of the table compares results at a rendering resolution of 384~$\times$~128, for comparison with the model of Tulsiani \etal. The lower part shows that our model also performs well when evaluated at full resolution. The `w/o background' rows show the result of taking our  MPI and ignoring the background image by replacing each layer's color with the input $\Image_s$. Especially on disoccluded pixels, using the background image leads to a substantial improvement. See~\citesec{sec:kitti}.}

\label{tab:kitti}
\end{table}

\begin{figure*}[t]
\centering
\includegraphics[width=0.98\textwidth,trim={.6em 0 0 0}]{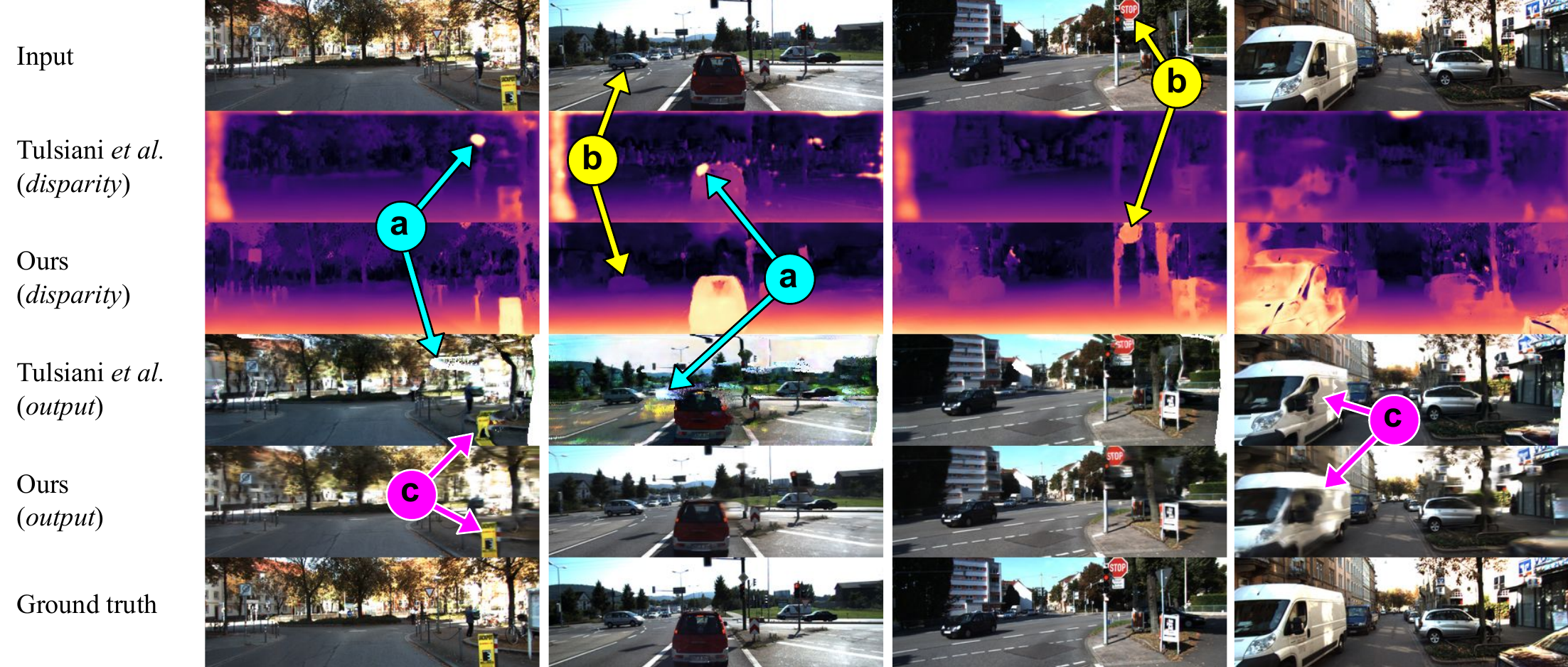}
\vspace{0pt}
\caption{Comparison on KITTI city sequences. (a)~Erroneous bright spots in Tulsiani \etal's depth-maps lead to unpleasant visual artefacts. (b)~Comparing depth maps shows structures (the car, the Stop sign) identified by our method but missing from theirs. (c)~In challenging areas their method produces sharp but distorted output, ours tends to produce blurrier output. See~\citesec{sec:kitti}.}
\label{fig:kitti_examples}
\end{figure*}

\subsection{View synthesis on Flowers light fields}
\label{sec:flowers}
We now apply our method to other datasets. Srinivasan \etal introduced a dataset of light field photos of flowers, and a method for predicting the entire light field from a single image \cite{srinivasan:2017:lightfield}. This dataset consists of over 3000 photographs, each one capturing a narrow-baseline 14 $\times$ 14 grid of light field angles. This dataset has no point cloud data for determining scale, so we cannot apply our scale-independent view synthesis approach. However, the scale is constant across the whole dataset so we can simply set $\sigma = 1$ and rely on the network to learn the appropriate scale. For this task, we add a gradient term to our synthesis loss. We train our model on the Flowers dataset by picking source and target images randomly from the 8~$\times$~8 central square of light field angles, and evaluate the results on a held-out set of 100 light fields. For comparison, we retrained the model of Srinivasan \etal \cite{srinivasan:2017:lightfield} using their publicly available code.

As shown in \citefig{fig:flowers}, our method improves on that of Srinivasan~\etal. Even if we eliminate the background image during testing (i.e.~set color $c_i = \Image_s$ throughout) we achieve higher PSNR and SSIM measures (along with lower absolute error) on the test set; using the predicted background image we see an additional small improvement. Our method has other advantages over theirs: we do not require complete light field data for training, and our representation can be rerendered at arbitrary novel viewpoints without further inference steps. \citefig{fig:flowers_examples} shows some qualitative comparisons.

\subsection{View Synthesis on KITTI}
\label{sec:kitti}
Instead of sampling source and target viewpoints from a sequence or light field, we can also apply our model to data where only left-right stereo pairs are available, such as KITTI \cite{geiger:2013:kitti}. Tulsiani \etal showed the possibility of using view synthesis as a proxy task to learn a two-layer layered depth image (LDI) representation from such data \cite{tulsiani:2018:lsi}. We train our model on the same data using 22 of the \emph{city} category sequences in the `raw' KITTI dataset, randomly taking either the left or the right image as the source (the other being the target) at each training step. Because the cameras are fixed, the relative pose is always a translation left or right by about 0.5m. Again, the scale is constant so we set $\scale = 1$, and again we add a gradient term to the synthesis loss. We compare our synthesized views on 1079 image pairs in the 4 test sequences with those produced by their pre-trained model. 

The representation used by Tulsiani \etal is less capable of high-quality view synthesis than our MPIs, because lacking alpha it cannot model soft edges, and because its splat-based rendering generates low-resolution output to avoid cracks. Both methods exhibit many artefacts at the image edges, so we crop 5\% of the image away at all sides, and then compute PSNR and SSIM metrics on all pixels, and also on `disoccluded' pixels only (as estimated by a multi-view stereo algorithm). For a fair comparison with Tulsiani \etal, our `low res' model matches theirs in resolution; we also train a `full' model at a higher resolution. Both models achieve improvements over theirs. The effect of our predicted background is small over the whole image, but larger when we consider only disoccluded pixels. Results are shown in \citetab{tab:kitti}, and qualitative comparisons in \citefig{fig:kitti_examples}.

\section{Conclusion}

We demonstrate the ability to predict MPIs for view synthesis from single image inputs without requiring ground truth 3D or depth, and we introduce a scale-invariant approach to view synthesis that allows us to train on data with scale ambiguity, such as that derived from online video. Our system is able to use the predicted background image to `inpaint' what lies behind the edges of foreground objects even though there is no explicit inpainting step in our system---although we typically do not see inpainting of more than a few pixels at present. A possible future direction would be to pair MPI prediction with adversarial losses to see if more, and more realistic, inpainting can be achieved.

\vfill\hrule\medskip\noindent{\small\textbf{Acknowledgements.}
This work benefited from helpful discussions with Jon Barron, Tali Dekel, John Flynn, Graham Fyffe, Angjoo Kanazawa, Andrew Liu and Vincent Sitzmann.
}

\newpage
{\small
\bibliographystyle{ieee_fullname}
\bibliography{bib}
}

\end{document}